\documentclass{ecai}
\usepackage{times}
\usepackage{graphicx}
\usepackage{latexsym}

\usepackage{cite}
\usepackage{amsmath,amssymb,amsfonts}
\usepackage{ tipa }
\usepackage{graphicx}
\usepackage{textcomp}
\usepackage{xcolor}
\usepackage{ esint }
\usepackage{graphicx}
\usepackage{mathtools}

\usepackage{algorithmic}
\usepackage[algoruled,linesnumbered]{algorithm2e}

\DeclarePairedDelimiter\ceil{\lceil}{\rceil}

\usepackage[thinlines]{easytable}
\usepackage{tabu}

\SetKwInput{KwInput}{Input}                % Set the Input
\SetKwInput{KwOutput}{Output}              % set the Output

\begin{document}

\title{Multi-Stage Transfer Learning with an
Application to Selection Process\\}

\author{Andre Mendes\institute{New York University, USA, email: andre.mendes@nyu.edu} 
\and Julian Togelius\institute{New York University, USA, email: julian.togelius@nyu.edu} 
\and Leandro dos Santos Coelho\institute{Pontifical Catholic University of Parana and Federal University of Parana, Brazil, 
email: leandro.coelho@pucpr.br}}

\maketitle

\begin{abstract}
In multi-stage processes, decisions happen in an ordered sequence of stages. Many of them have the structure of dual funnel problem: as the sample size decreases from one stage to the other, the information increases. A related example is a selection process, where applicants apply for a position, prize or grant. In each stage, more applicants are evaluated and filtered out and from the remaining ones, more information is collected. In the last stage, decision-makers use all available information to make their final decision. To train a classifier for each stage becomes impracticable as they can underfit due to the low dimensionality in early stages or overfit due to the small sample size in the latter stages. In this work, we proposed a \textit{Multi-StaGe Transfer Learning} (MSGTL) approach that uses knowledge from simple classifiers trained in early stages to improve the performance of classifiers in the latter stages. By transferring weights from simpler neural networks trained in larger datasets, we able to fine-tune more complex neural networks in the latter stages without overfitting due to the small sample size. We show that is possible to control the trade-off between conserving knowledge and fine-tuning using a simple probabilistic map. Experiments using real-world data show the efficacy of our approach as it outperforms other state-of-the-art methods for transfer learning and regularization.
\end{abstract}

\section{Introduction}
%---------------------------------------------------------------------------------------
In many practical applications including homeland security, medical diagnosis, and network intrusion detection, decision systems are composed of an ordered sequence of stages~\cite{trapeznikov2012multi}. This can be referred to as a multi-stage process. To train classifiers to work in such processes, many aspects have to be observed, since they present different structures and the level of decision, data and sample size can significantly change in each stage.

One specific common example of such a process is a selection process. A selection process can be defined as the process of selection and shortlisting of the right candidates with the necessary qualifications and skill set to fill determined positions~\cite{marsden1994hiring}. Although the majority of examples are related to job opportunities, selection processes are also used for fellowships, grants, and prizes. 

The simplest type of selection process is comprised of one single stage where the evaluator has information about the applicant and makes a decision. However, the more elaborated the process is, the more stages are necessary to filter the applicants, as not everyone is given the same attention to save resources.

\begin{figure}[tbp]
\centerline{\includegraphics[width=\columnwidth]{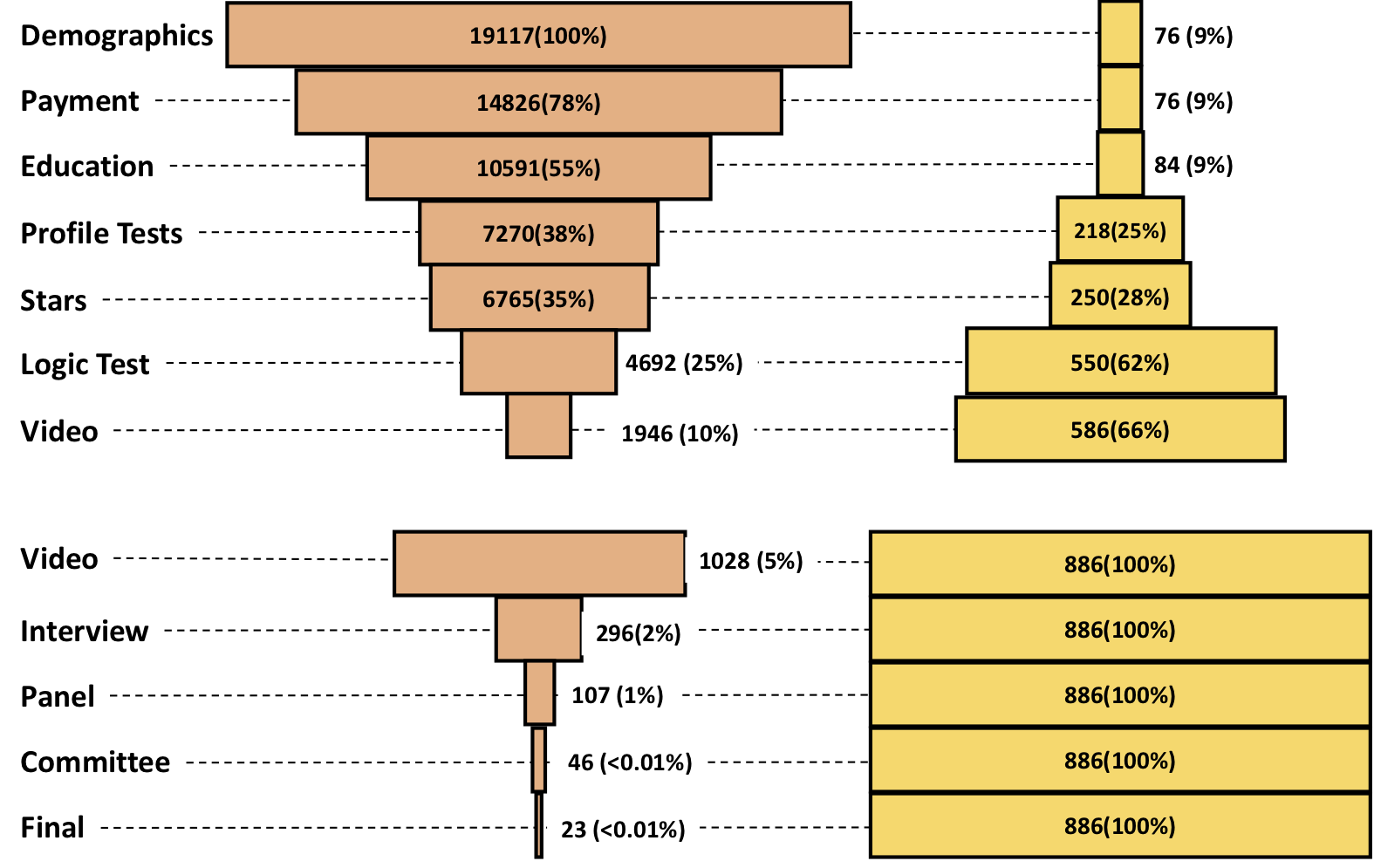}}
\caption{Representation of a dual funnel problem. The left funnel shows the number of applicants decreasing, whereas the right funnel shows the amount of data (in terms of variables) increasing during the process. The top and down parts show the stages in the conversion and evaluation phase, respectively.}
\label{fig:inverse_funnel}
\end{figure}

One of the hardest challenges for training classifiers for a multi-stage process can be defined as a dual funnel problem (see Figure \ref{fig:inverse_funnel} for details). During each stage of the conversion phase, the number of applicants decrease whereas the data about them increases, which means that the dataset grows in dimensionality. 

The classifiers that are trained in initial stages have sufficiently large samples to generalize. However, there might not be useful information to differentiate between the best applicants, so the models have high bias and tend to underfit. In the final stages, there is plenty of information for each applicant, but the sample size is so low that classifiers trained only on this stage suffer from high variance and tend to overfit the data.
%---------------------------------------------------------------------------------------
%---------------------------------------------------------------------------------------

Here we address this problem using \textit{Transfer Learning} (TL)~\cite{pan2010a,weiss2016survey}. In TL, the goal is to use the knowledge obtained from solving a problem in one domain to solve a related problem in a different domain. We apply this methodology here to transfer knowledge learned in each stage to a subsequent stage to create better classifiers. 

In our approach, called \textit{Multi-StaGe Transfer Learning} (MSGTL), we propose a structure where simpler and more general \textit{neural networks} (NN) are trained in the initial stages. Such NNs are more general because they were trained in a dataset with a large sample size but small dimensionality. In later stages, the weights from this NN will be transferred to more complex NNs trained in datasets with more features but a much smaller sample size. The general knowledge transferred from the early stages helps the more complex NN to prevent overfitting and achieve better performance than a classifier trained in the single-stage only. We show that our approach can overcome the common problems in transfer learning and classical challenges from real data such as covariate shift and class imbalance.

The main contributions of this paper are:
\begin{enumerate}
    \item We present a framework that allows knowledge to be transferred between NN structures in different stages in a multi-stage problem. Our approach can handle different types of unstructured (text) and structured data (tabular, categorical).
    \item We introduce a hyperparameter to control the knowledge transferred during training in each stage. We show that our hyperparameter can be used to find an appropriate trade-off between keeping the knowledge to prevent overfitting and allowing fine-tuning to improve performance.
    \item Our method improves on cascaded classifiers and other multi-stage structures by making full decisions in every stage with available data.
    \item The efficacy of the proposed model is demonstrated by extensive longitudinal and cross-validated experiments on the real data from a selection process in 3 distinct years. We show that our model can outperform other state-of-the-art methods in transfer learning and regularization in the multi-stage case. Our method presents significant gains particularly in the late stages of the process with smaller sample sizes.
\end{enumerate}

The remainder of this paper is organized as follows. Section~\ref{sec:related_work} provides an overview of the related work in the field. Section~\ref{sec:problem_statement} defines the problem and in Section~\ref{sec:MSGTL}, the proposed method is presented. Section~\ref{sec:experiments} explains the experiments used to validate the method and the results are shown in Section~\ref{sec:results}. Finally, Section~\ref{sec:conclusion} presents the conclusion.
%---------------------------------------------------------------------------------------
\section{Related Work}
\label{sec:related_work}
%---------------------------------------------------------------------------------------
The work presented here is most closely related to two subfields of machine learning, namely transfer learning~\cite{pan2010a,weiss2016survey} and multi-stage classifiers~\cite{senator2005multi} or decision cascades~\cite{saberian2010boosting}. Additionally, our method is similar to some regularization techniques, such as dropout.
%---------------------------------------------------------------------------------------
\subsection{Transfer Learning}
%---------------------------------------------------------------------------------------
\textit{Transfer Learning} (TL) focuses on storing knowledge gained while solving one problem and applying it to a different but related problem. Humans do this process in many cases such as using knowledge from playing tennis to play table tennis.

Following the definitions presented in~\cite{weiss2016survey}, TL can be defined in two categories. The first one is called homogeneous transfer learning and it refers to problems where the feature spaces in both domains are the same. The main focus is on aligning data distributions between domains through domain adaptation. Previous attempts to solve this problem have achieved important results ~\cite{pan2010domain,zhang2013domain}, especially in computer vision~\cite{gong2012geodesic} and natural language processing~\cite{collobert2011natural}. The second case is called heterogeneous transfer learning and the main problem is to align feature spaces between domains to achieve representation transformation. There are mainly two approaches used in this case: map $X_T$ to $X_S$, which is referred to as asymmetric transformation,~\cite{feuz2015transfer} or map both $X_S$,$X_T$ to a
common latent space, symmetric transformation~\cite{duan2012learning}. 

Other methods use \textit{deep neural networks} (DNN) that can better learn abstract representations of the data on different levels and generality~\cite{bengio2013representation,yosinski2014transferable}. With this ability, DNNs have been used to achieve state-of-the-art results in many transfer learning problems~\cite{guo2016deep,rozantsev2018beyond}. Additionally, adversarial learning has been successfully explored to minimize cross-domain discrepancy. For example, a simple and efficient gradient reversal layer is proposed in~\cite{ganin2016domain} to promote the emergence of features that are discriminative for the main learning task on the source and indiscriminate concerning the difference in the stages. Other methods have extended this approach for the multi-domain case~\cite{long2018conditional,liu2019transferable}.

Many methods rely on weight transfer, where a NN is pretrained on a source task and the weights of some of its layers are transferred to a second NN that is used for another task. The non-transferred weights of the NN are randomly initialized and a second training phase follows~\cite{yosinski-a,azizpour2015a}.
%----------------------------------------------------------------------------
\subsection{Multi-Stage classification}
\begin{figure*}[tbp]
\centerline{\includegraphics[width=1\linewidth]{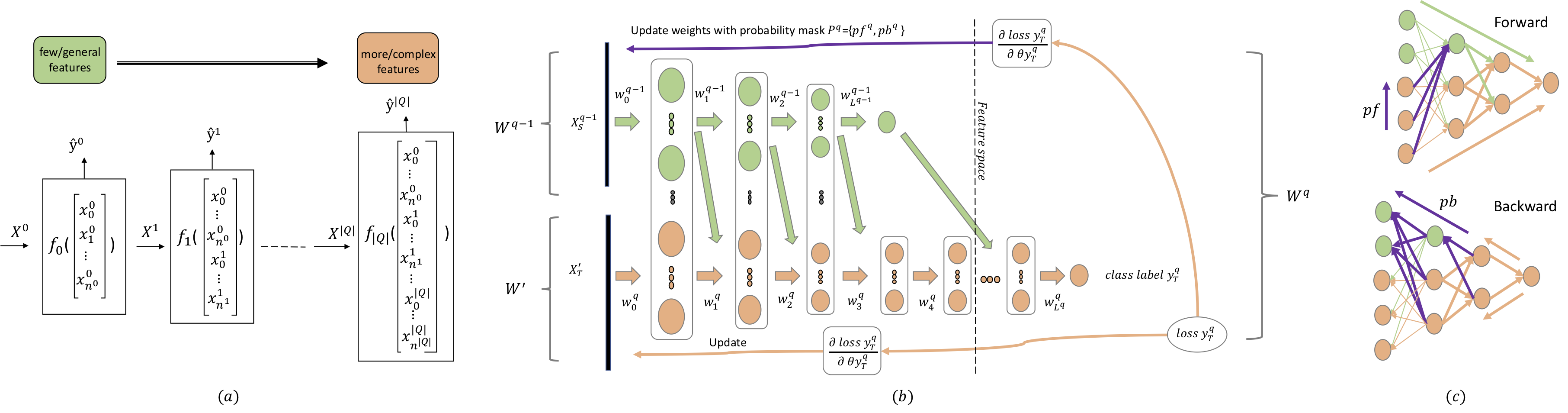}}
\caption{Representation of a multi-stage architecture (a) and the proposed approach for weight transfer (b-c). For every stage $q \in Q$, a set of features $X^q=X^{q-1}||X'$ is available (the symbol $||$ represents the column-wise concatenation of two matrices). In (a), a classifier $f_q$ makes the decisions $\hat{y}^q$. Our proposed method is shown in (b), where the weights from the \textit{neural network} (NN) structure in $W^{q-1}$ (green) are transferred and together with the new weights in $W'$ (orange), they form the new NN structure $W^q$ (green+orange). The orange weights are initialized randomly and are updated during training. The green weights are initialized using $W^{q-1}$ and their updates are controlled using binary masks defined by the probabilistic variable $\rho$. In (c), the top part shows the updates in the forward pass controlled by $pf^q$. The bottom part shows the backward pass controlled by $pb^q$. A purple weight will only be updated if its correspondent value in the binary mask is 1; otherwise it stays frozen.}
\label{fig:weight_transfer}
\end{figure*}

%----------------------------------------------------------------------------
Multi-stage and cascade classifiers share many similarities. However, an important difference between cascade~\cite{viola2001robust} and multi-stage can be defined as the system architecture. Detection cascades make partial decisions, delaying a positive decision until the final stage. In contrast, multi-stage classifiers can deal with multi-class problems and can or have to make classification decisions at any stage~\cite{trapeznikov2012multi}. In our case, our method needs to make a full decision in each stage, since there will only be more features in the next one. Figure~\ref{fig:weight_transfer}(a) shows an example of a multi-stage architecture.
%----------------------------------------------------------------------------
%----------------------------------------------------------------------------

Cascade classifiers are an important part of a subarea of machine learning that researches the trade-off between learning and feature acquisition or learning with budget constraints~\cite{trapeznikov2012multi,trapeznikov2013supervised}. It has also been used for content classification~\cite{alfaro2016multi} and detecting instructions and anomalies in networks~\cite{cordella2007multi}. However, the vast applications recently have been to unstructured data in computer vision, in areas such as object segmentation~\cite{luo2012hierarchical}, scene understanding~\cite{ranzato2011deep}, and recognition~\cite{jarrett2009best}. We focus on structured data in tabular form with categorical attributes and unstructured data in text format.

The approaches in ~\cite{zeng2013multi,sabokrou2017deep,qi2019tourism} explore the connection between deep models and multi-stage classifiers such that classifiers can be jointly optimized and they can cooperate across the stages. This structure is similar to our approach. However,  our algorithm only has access to the information in a specific stage, whereas in the other methods, the entire dataset is available and a decision is made about which parts to use.

To avoid losing information from one stage to another, cascades can use soft classifiers, which collects the classification scores extracted by each stage of classifiers and then combines the classification scores for the final decision~\cite{dollar2012crosstalk}. Similarly, we use the classification scores from the classifiers in each stage to inform the next NN in our structure. However, these classifiers are not soft and don't have a reject mechanism, so the scores come from the actual decisions in each stage.
%----------------------------------------------------------------------------
\subsection{Probabilistic Regularization}
%----------------------------------------------------------------------------
Regularization is an important part of training NN to improve generalization capabilities. Dropout~\cite{srivastava2014dropout} is a technique that improves the generalization of NN by preventing the co-adaptation of feature detectors. The working of dropout is based on the generation of a mask by utilizing Bernoulli and Normal distributions.

Following the success of Dropout, other methods have been explored different aspects, such as sampling dropout mask from different distributions ~\cite{kingma2015variational,gal2016dropout} or actively selecting nodes based on their strength in the NN~\cite{keshari2019guided}. Similar to these methods, our approach also uses a binary mask to control updates in the weights. However, a key difference is that we preserve the knowledge acquired in previous states, by using the mask only in new nodes introduced in later stages. Additionally, our mask contains a forward and backward component that controls the active state of a node in both forward and backward passes through the NN.
%----------------
\section{Problem Statement}
\label{sec:problem_statement}
%------------------------------
A domain $\mathcal{D}$ is defined by a feature space $\mathcal{X}$ and a marginal probability distribution $P(X)$. Each $x_i$ is the i-th feature vector (instance), $n$ is the number of feature vectors and $m$ is the number of samples in $X \in \mathbb{R}^{m{\times}n}$. We need a predictive function $f(\cdot)$ from the feature vector and label pairs $\{x_i, y_i\}$ where $x_i \in X$ and $y_i\in\mathcal{Y}$. From the definitions above, a domain $\mathcal{D}=\{\mathcal{X},P(X)\}$ and a task $\mathcal{T}=\{\mathcal{Y},f(\cdot)\}$ are defined. Therefore, $\mathcal{D}_S$ and $\mathcal{T}_S$ are the source domain and task, respectively. Similarly, $\mathcal{D}_T$ and $\mathcal{T}_T$ are the target and task respectively. We can use TL to improve the target predictive function $f_{T}(\cdot)$ by using the related information from $\mathcal{D}_S$ and $\mathcal{T}_S$.

Let’s define $Q$ as the stage vector and a single stage $q \in Q$. In a multi-stage process, we can define previous stages as the source domain $\mathcal{D}_S$, and the following stage as the target domain $\mathcal{D}_T$.

For each stage in a dual funnel problem, new data is received while the number of samples decreases, which means $X^{q-1}\neq X^{q}$, $m^{q-1}>m^{q},n^{q-1}<n^{q}$. As $m$ gets significantly smaller in absolute value and in comparison to $n$, classifiers trained on the specific stage data tend to overfit. With new features in a new stage, we also have a shift in the feature space, $\mathcal{X}_S\neq\mathcal{X}_T$. This requires that a domain adaptation so that the features in the source in the target can be in the same feature space.

The process also contains domain class imbalance, defined as $P_S(Y)\neq P_T(Y)$. This happens because the distribution between positives and negatives samples is different across source and target. This is a challenge for many transfer learning algorithms~\cite{weiss2016investigating}.

With all together, our method tries to solve the problem of transferring knowledge from one stage to the other in order to address overfitting caused by $m^{q-1}>m^{q},n^{q-1}<n^{q}$, the shift in the feature space $\mathcal{X}_S\neq\mathcal{X}_T$, and the domain class imbalance $P_S(Y)\neq P_T(Y)$. Our method does not directly address domain class imbalance, but we use different penalties for misclassifications on positive samples to address the problem of class imbalance in each stage.
%---------------------------------------------------------------------------------------
\section{Multi-Stage Transfer Learning (MSGTL)}
\label{sec:MSGTL}
%---------------------------------------------------------------------------------------
Our proposed method uses transfers weights to share knowledge between the stages. Figure \ref{fig:weight_transfer} shows the main idea for our approach. To address overfitting, we propose a simple weight transfer mechanism to use knowledge from previous stages to train the NN in later stages. The weights trained in the initial stages are exposed to larger and simpler data. By transferring them to the later stages, we add additional regularization for the more complex structures.

In the first step, our method learns a simpler NN that is trained in a dataset with more samples and less complexity. Later, the weights from this NN will be used as initialization for a more complex NN trained in a dataset with fewer samples but more features. During the training of the second NN, the weights from the previous one are updated according to the respective probabilistic map. The knowledge shared between NNs allows the second to perform better in the more complex dataset without overfitting due to the smaller sample size. 

For the shift in feature space, we augment our NN structure with a gradient reversal layer proposed in~\cite{ganin2016domain}. This method uses adversarial learning where a domain discriminator is trained to distinguish the source from the target and the feature representation is trained simultaneously to confuse the domain discriminator, decreasing the gaps between the feature spaces. To improve the discriminator, we add a weighted loss function where the weights represent the difference in the number of samples from source and target.

In our method, the NN is trained sequentially incorporating information from the entire process. It is important to note that our approach is different than an ensemble, in which a group of classifiers is trained separately, with no information shared, and the decisions are typically made based on a voting system. In our approach, the weights from previous stages are combined with new ones, affecting the batch update and sharing information in all layers. 
Our method also differs from other weight transferring approaches. Traditional techniques transfer the entire layer of the NN, usually the early ones with general attributes and fine-tune the last layers with the new data. In our method, we combine transferred and new weights in all layers and control the fine-tuning with a probabilistic variable.
%---------------------------------------------------------------------------------------
\subsection{Architecture}
%---------------------------------------------------------------------------------------
The components of the proposed method are:
%--------------------------------------------------------------------------------------
\begin{itemize}
    \item We redefine $\mathcal{X}$ and $\mathcal{Y}$ as vectors containing all input features and labels, respectively for the entire process.
    \item $q \in Q$ represents a single stage in the process with $|Q|$ stages.
    \item $X^q \in \mathbb{R}^{m^{q}\times n^q}$ is the feature matrix input and $Y^q \in \mathbb{R}^{m^{q}\times 1}$ are the labels for current stage $q$.
    \item $h^q_l$ is the number of nodes in a given layer $l=\{0,1,...,L^q\}$.
    \item $L^q$ represents the number of layers in the NN. To simplify our structure, we start the NN with $n^q$ nodes and each following layer has half of the nodes compared to the previous one. Therefore, we have $h_l^q=2*h^q_{l+1}$ for $l=\{0,..L^q\}$. We define a max number of layers $\omega$ to bound the NN in case the initial input feature vector is too large (analysis of the hyperparameters in section \ref{sec:parameter_sensitivity}). Finally, we also define a hyperparameter $\gamma$ for the number of nodes in the last activation layer before the final classification. Therefore:
    \begin{equation}
        L^q =
        \begin{cases}
            3, & \textit{if} \ n^q\leq\gamma \\
            \omega, & \textit{if} \ \ceil{\log_2(\frac{n^q}{\gamma})}+2 \geq \omega\\
            \ceil{\log_2(\frac{n^q}{\gamma})}+2, & \textit{otherwise}
        \end{cases}
    \end{equation}
    \item $W^q=\{w_0^q,w_1^q,...w_{L^q}^q\}$ is a vector containing the NN structure for stage $q$.
    \item $w_l^q \in \mathbb{R}^{h^q_l\times h^q_{l+1}}$ is a single weight matrix that connects two consecutive layers in the NN.
    \item $P^q=\{p_0^q,p_1^q,...p_{L^q}^q\}$ is a vector containing all binary masks corresponding to parameters in $W^q$. 
    \item $p_l^q=\{pf_l,pb_l\}$ is formed by the forward mask $pf_l$, and the backward mask $pb_l$. Each component is a binary matrix with independent random variables drawn from a Bernoulli distribution with probability $\rho$ of being 1.
    \item $\mathcal{W} = \{W^0,W^1,...,W^{|Q|}\}$ is the final vector containing all the NN structures for the entire process.
\end{itemize}
%--------------------------------------------------------------------------------------
Considering a standard neural network structure, our modifications with $\rho$ and $P$ can be result in a loss function similar to the case in Dropout~\cite{srivastava2014dropout} and it can be solved with standard error backpropagation method. Additionally, we address the problem with class imbalance using the balanced cross-entropy loss function for optimization defined as 
\begin{equation}
    \mathcal{L}=-\sum_{i=1}^{m} \beta y_{i} \log \hat{y}_{i}+(1-\beta)(1-y_{i})\log(1-\hat{y}_{i}),
\end{equation}
where $y$ represents a ground truth label, $\hat{y}$ is a predicted label and $\beta$ is the fraction of the sample which is dominant in a dataset, $\beta=1-\frac{\sum_{y \in Y}}{|Y|}$ when $y \in \{0,1\}$. This modification helps the model to focus on learning positive labels, since the learning for the dominant negative class is reduced.
%--------------------------------------------------------------------------------------
\subsection{Step-by-Step training}
%--------------------------------------------------------------------------------------
The algorithm for our method is presented in Alg.~\ref{alg:MSGTL}. Let's define the general case with $start=0$. In $q=0$ a neural network receives input features and labels from $X^{0}$ and $Y^{0}$. The first layer of the NN receives a input feature vector from $X^{0}=\{x^0_0,...,x^0_{n^0}\}$ and all other parameters, $\{w^0_0,...,w^0_L\}$ are randomly initialized. The masks in $P^0$ are defined with all values equal to 1. We train the initial NN, creating the NN structure $W^0=\{w^0_0,...,w^0_L\}$.
    
In the following stage $q$, features and labels come from $X^q$ and $Y^q$, respectively. We first initialize a new NN structure $W^q$ with random weights. Then in line 10, we loop through the parameters in $W^{q-1}$ updating the respective weights in $W^q$ so that the pretrained values in $W^{q-1}$ are passed to $W^q$. In the same process, we also create the masks $P^q=\{pf_l^q,pb_l^q\}^{L^q}_{l=0}$, where $pf^q_l$ and $pb^q_l$ are the binary masks defined by $Bernoulli(\rho)$.

During training, all the weights that were initialized randomly are updated. As for the pretrained weights, they will be updated according to the binary masks $pf^q_l$ and $pb^q_l$ during the forward and backward pass respectively. If $\rho=0$, all the pretrained weights are kept frozen. We call this \textit{fixed configuration}. In the case of $\rho=1$, all weights are updated. We call this \textit{initialization configuration} since the pretrained weights are only initialized with the previous structure but they are fine-tuned during training (more details for $\rho$ in \ref{sec:parameter_sensitivity}). 

After all the stages are completed, we have a vector $\mathcal{W}$ that contains a neural network for each stage. For inference, each NN is retrieved and used to make the predictions for the respective stage.
%--------------------------------------------------------------------------------------
\begin{algorithm}[tb]
\caption{Train MSGTL}
\label{alg:MSGTL}
\textbf{Input}: $x_{all},y_{all},\omega,S,pu,\lambda,start,epochs$\\
\textbf{Output}: Optimized $\mathcal{W}$
\begin{algorithmic}[1] %[1] enables line numbers
\STATE $\mathcal{W}\leftarrow\{\}$; $s\leftarrow start$
\STATE $X^s,y^s\leftarrow \mathcal{X}[s],\mathcal{y}[s]$
\STATE $W^s\leftarrow Initialize(W,\gamma,\omega)$
\STATE $P^s\leftarrow Ones(W^s)$ 
\STATE $W^s\leftarrow Train(X^s,y^s,W^s,P^s,\lambda,epochs)$
\STATE Add $W^s$ in $\mathcal{W}$
    \FOR{$s=start+1$ to $S$}
        \STATE $X^s,Y^s\leftarrow x_{all}[s],y_{all}[s]$
        \STATE $W^s\leftarrow Initialize(W,\gamma,\omega)$
        \STATE $W^s,P^s \leftarrow TransferWeights(W^{s-1},W^{s},\rho)$
        \STATE $W^s\leftarrow Train(X^s,y^s,W^s,P^s,\lambda,epochs)$
        \STATE Add $W^s$ in $\mathcal{W}$
    \ENDFOR
\end{algorithmic}
\end{algorithm}
%---------------------------------------------------------------------------------------
\section{Experiments}
\label{sec:experiments}
\begin{table*}[tbp]
\caption{Stages in the multi-stage selection process}
\label{tab:process_description}
\scriptsize
%\begin{tabu}{|c|p{0.85\linewidth}|}
\tabulinesep=2pt
\begin{tabu}{|>{\centering\arraybackslash}m{.1\linewidth}|m{.85\linewidth}|} 
\hline
\multicolumn{2}{|c|}{Conversion Phase} \\  \hline
Demographics & Provide information about the country, state, city, and age. The system performs automatic validations to see if they match the criteria for the process. \\  \hline
Payment & Pay the application fee or ask for a waiver. For a waiver, the applicant needs to write a justification. \\ [5pt] \hline 
Education & Provide information on the university, major and extra activities. \\  \hline
Profile Tests & Perform online tests to map profile characteristics such as interests, values, mindset, and personality. \\  \hline
Star & Write about life experiences that were meaningful following a STAR structure (S: situation, T: task, A: action, R: result). \\  \hline
Logic Tests & Respond to online tests to map levels in problem-solving involving logic puzzles. \\  \hline
Video Submission & Submit a 2-min long video, explaining why they deserve the fellowship. There are no restrictions on the content of the video and the quality of the recording. \\ \hline
\multicolumn{2}{|c|}{Evaluation Phase} \\  \hline
Video Evaluation & Applicants are evaluated based on their entire profile submitted. This is the first stage where students are accepted or rejected by evaluators. \\  \hline
Interview & Applicants have the first direct contact with the evaluators. In this stage, the goal is to have a screening interview where the evaluators can go deeper and clarify questions about the applicant's profile. \\  \hline
Panel & Former fellows are evaluators in a panel section interview. The panel has 3 to 4 fellows that interview 5 to 6 applicants at the same time. \\  \hline
Committee & The selection team and some of the senior fellows discuss the results and select applicants to move to the final step. Applicants do not participate in this step. \\  \hline
Final & Applicants are interviewed by the direct board of the foundation. The interview structure is similar to the panel structure and the applicants are evaluated together. This is the final stage where the applicants will be selected for the fellowship. \\  \hline
\end{tabu}
\end{table*}
%--------------------------------------------------------------------------------------
To validate our method we perform a series of experiments using a real-world dataset from a multi-stage selection process. This section presents the datasets, feature preparation, performance metrics, validation methods and the algorithms that we use for a benchmark. 
%--------------------------------------------------------------------------
\subsection{Datasets}
%--------------------------------------------------------------------------
Our dataset comes from an organization that select students for a fellowship. This selection process contains two phases, conversion and evaluation as shown in Figure \ref{fig:inverse_funnel} and Table \ref{tab:process_description}. In conversion, the applicants apply for the fellowship and fill up general information in a web portal. For each new stage, the applicants enter new and more detailed information. In this phase, there is no interference from humans evaluators and the applicants can stop the process, or be filtered out by automatic filters or conditions. All the applicants that finish the first phase move to the second phase called evaluation. In this phase, the applicants are selected based on the criteria for each stage. Fewer applicants continue in the stages until the final selection. 

With regards to data, the video stage marks the last part where data about the applicants is collected in a structured, objective and automatic form. In the following stages that require interviews, even though the applicants present more information, only the evaluators have access to it. All decisions in all stages are binary, meaning if the applicant was accepted or rejected.
%--------------------------------------------------------------------------------------
\subsection{Feature Preparation}
%--------------------------------------------------------------------------------------
In the first stages, all the data obtained from the applicants is structured in a tabular format. For categorical data, such as the state where they live, a standard one hot encode transformation is used to convert the data to a numerical form. For more complex datasets such as education, there is a feature engineering step to reduce the dimension size of the dataset. This includes, for example, creating groups for universities and excluding strong correlated features. For the datasets from profile or logic tests, all the information is already in continuous numerical form and we perform a standard normalization.

In the later stages of the process, such as stars and video, the data is unstructured. The stars are open answers in form of text. For the video, we only consider the content of the speech component. The speech is extracted and used as a text. To convert this unstructured data to numerical values, we create word embeddings using Word2Vec~\cite{mikolov2013efficient}. The goal is to assign high-dimensional vectors (embeddings) to words in a text corpus in a way that preserves their syntactic and semantic relationships. To train the embeddings, we use a pretrained model trained on the entire Wikipedia in Portuguese from Brazil and extracted a 300-dimensional vector representation for each word. We tested different methods to aggregate the words and the best method was standard average in the word vectors.
%--------------------------------------------------------------------------------------
\subsection{Validation and Metrics}
%--------------------------------------------------------------------------------------
For the experiments, we first perform cross-validation using each dataset for each year separately. For example, the dataset from applicants in 2017 is split into training, test, and validation. In the second batch, we combine pairs of years, such as 2017 and 2018, and perform the random cross-validation split in the aggregated data. Finally, we perform the same experiment using data from all years combined. The combinations result in 7 experiments using 10-fold cross-validation in each one of them for a total of 70 runs.

In the longitudinal experiments, we use a previous year as a training and test set and the following year as a validation set. For example, we split the dataset from 2017 and train and test, find the best model and validate its results using the dataset from 2018. The model, in this case, is trained in 2017 and has never seen any data in 2018. We repeat the process for 2017 and 2019, and 2018 and 2019. Finally, we also combine the datasets between 2017 and 2018 and validate the results in 2019. Therefore, we have 4 different experiments with 10-fold cross-validation resulting in a total of 40 runs.

Since our dataset has a funnel structure, it is common in this scenario to have an imbalanced dataset. Moreover, in a selection process, there is also a significant difference between misclassifying a weaker applicant as strong (false positive) then a misclassifying a stronger applicant as weak (false negative). Therefore, we choose to compare the models in terms of f1-score, which balances precision and recall, for the positive class.
%--------------------------------------------------------------------------
\begin{figure*}[tbp]
\centerline{\includegraphics[width=.9\linewidth]{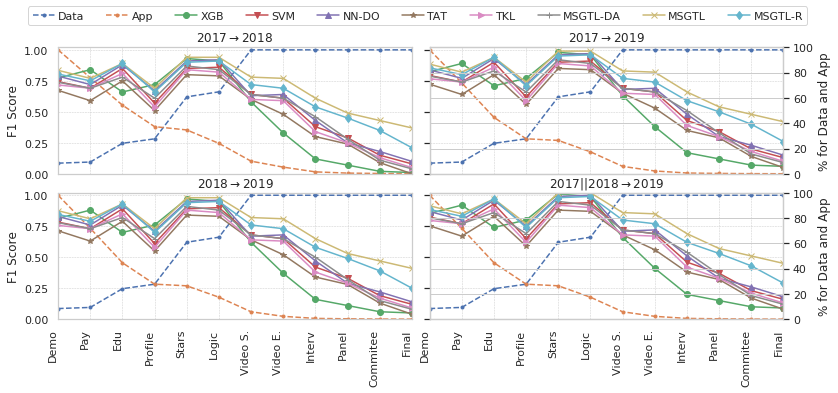}}
\caption{Results obtained in the longitudinal setup. For each new stage, the number of applicants drops while more data about them is obtained. Algorithms perform better while the number of samples is relatively high but the performance worsens drastically in the later stages. The percentage of applicants and data in each stage is shown on the y-axis. For algorithms, the y-axis represents the performance in terms of F-score for positive class.}
\label{fig:results_lineplot}
\end{figure*}
%--------------------------------------------------------------------------
\subsection{Benchmark Methods}
%--------------------------------------------------------------------------
We compare the results with other established methods, following the choices for optimization and regularization defined in the experiments for each paper and for the selection of the hyperparameters, we conduct \textit{importance-weighted cross-validation} (IWCV)~\cite{sugiyama2007covariate}. 

Due to the size of the dataset, especially in the last stages, we expect that commonly used non-deep algorithms such as \textit{Logistic Regression} (LR)~\cite{hosmer2013applied} and \textit{Support Vector Machines} (SVM)~\cite{chang2011libsvm} would achieve better performance because tend to generalize better on smaller datasets. The trade-off hyperparameter $C$ for both SVM and LR is $C=0.1$. We also compare our method with ensemble approaches such as \textit{Random Forest} (RF)~\cite{liaw2002classification} and \textit{XBoost} (XB)~\cite{chen2016xgboost} using trees with a maximum depth equals 8, shrinkage equals 0.1 and no column sub-sampling.

For domain adaptation methods, we use \textit{Domain Adversarial Neural Network} (DANN)~\cite{ganin2016domain}, \textit{Transferable Adversarial Training (TAT)}~\cite{liu2019transferable}, and \textit{Transfer Kernel Learning (TKL)}~\cite{long2014domain} with an SVM as the base classifier. For DANN and TAT, we follow the hyperparameters used in ~\cite{liu2019transferable} adopting Adam~\cite{kingma2014adam} with a inverse-decay strategy~\cite{ganin2016domain} where the learning rate changes by $\eta_p=\frac{\eta_0}{(1+\omega_p)^\phi}$, $\omega=10, \phi=0.75$, and $p$ is the progress ranging from 0 to 1. For TKL we use $\sigma=0.1, \lambda=10, p=10$ and $\gamma=10$. For each domain adaptation method, we perform training using two stages: the previous stage as the source domain and the current stage as the target domain.

For standard NN, we use regularization methods to counteract the tendency to overfit on small data. Specifically, we include the \textit{Virtual Adversarial Training (NN-VAT)}~\cite{miyato2018virtual} with hyperparameters $K=1, \epsilon=2$ and $\alpha=1$ and \textit{Dropout} (NN-DO)~\cite{srivastava2014dropout} with $p=0.5$. 

As for the method proposed in this work, we include tree variations: MSGTL is the base configuration with transfer weights from one stage to the other following different values of $\rho$. MSGTL-DA is the configuration where we add a domain adaptation component using the reverse gradient layer following the structure proposed in TAT~\cite{liu2019transferable}. The final variation is MSGTL-R, which combines the method proposed here with an extra regularization term from DO. We use TAT and DO because they performed better than other methods such as DANN and VAT, respectively. The best results in our experiments were achieving using $\omega=6, \gamma=2$ and $\rho=0.3$. 
%---------------------------------------------------------------------------------------
\section{Results}
\label{sec:results}
%---------------------------------------------------------------------------------------
In this section, we present the results obtained in the experiments described in section ~\ref{sec:experiments}. We also present an analysis of different values of the hyperparameters and how they affect the final performance.
%---------------------------------------------------------------------------------------
 \subsection{Longitudinal Validation}
%---------------------------------------------------------------------------------------
Table \ref{tab:stage_results} shows the average results obtained for each stage in all experiments. For each one of the groups, we show the results for the best algorithm: standard (SVM), ensemble (XB), neural networks with dropout (NN-DO) and domain adaption (TAT, TKL). From these results we can make the following observations:
\begin{table*}[tbp]
\caption{Average results for all methods in each stage using F1-Score for the positive class. Numbers in the left in each cell show results for the longitudinal setup while numbers in parenthesis represent results obtained in the cross-validation setup.}
\centering
\scriptsize
\tabulinesep=2pt
\label{tab:stage_results}
%\begin{tabular}{|l|l|l|l|l|l|l|l|l|l|}
\begin{tabu}{|c|l|l|l|l|l|l|l|l|l|} 
\hline
Steps & XGB & SVM & NN-DO & TAT & TKL & MSGTL-DA & MSGTL & MSGTL-R & \textbf{MEAN} \\ \hline
\multicolumn{10}{|c|}{Conversion Phase} \\ \hline
Demographics & \textbf{0.80  (0.81)} & 0.77  (0.79) & \textbf{0.8  (0.82)} & 0.70  (0.71) & 0.73  (0.73) & 0.75  (0.77) & \textbf{0.86  (0.89)} & \textbf{0.83  (0.85)} &  0.78  (0.80)\\ \hline
Payment & \textbf{0.85  (0.85)} & 0.7  (0.75) & 0.73  (0.77) & 0.6  (0.63) & 0.7  (0.72) & 0.7  (0.75) & \textbf{0.78  (0.83)} & 0.76  (0.79) & 0.73  (0.76) \\ \hline
Education & 0.67  (0.7) & 0.86  (0.87) & \textbf{0.89  (0.89)} & 0.76  (0.79) & 0.82  (0.85) & 0.79  (0.79) & \textbf{0.9  (0.92)} & \textbf{0.9  (0.94)} & 0.82  (0.85) \\ \hline
Profile & \textbf{0.73  (0.75)} & 0.58  (0.61) & 0.68  (0.73) & 0.52  (0.54) & 0.55  (0.55) & 0.62  (0.67) & \textbf{0.7  (0.75)} & 0.67  (0.69) & 0.63  (0.66) \\ \hline
Star & \textbf{0.94  (0.97)} & 0.86  (0.9) & \textbf{0.92  (0.94)} & 0.81  (0.83) & 0.85  (0.86) & 0.88  (0.89) & \textbf{0.95  (0.97)} & \textbf{0.91  (0.92)} & 0.89  (0.91) \\ \hline
Logic & \textbf{0.92  (0.93)} & 0.87  (0.91) & \textbf{0.93  (0.96)} & 0.8  (0.81) & 0.83  (0.88) & 0.85  (0.87) & \textbf{0.95  (0.97)} & \textbf{0.92  (0.96)} & 0.88  (0.91) \\ \hline
Video & 0.59  (0.63) & 0.65  (0.67) & 0.64  (0.64) & 0.61  (0.66) & 0.61  (0.63) & 0.65  (0.67) & \textbf{0.79  (0.8)} & \textbf{0.73  (0.74)} & 0.66  (0.68) \\ \hline
\textbf{MEAN} & \textbf{0.78  (0.81)} & 0.75  (0.79) & \textbf{0.8  (0.82)} & 0.68  (0.71) & 0.73  (0.73) & 0.75  (0.77) & \textbf{0.85  (0.88)} & \textbf{0.82  (0.84)} &  \\ \hline
\multicolumn{10}{|c|}{Evaluation Phase} \\ \hline
Video & 0.34  (0.42) & 0.62  (0.69) & 0.65  (0.72) & 0.49  (0.54) & 0.6  (0.67) & 0.62  (0.7) & \textbf{0.78  (0.81)} & \textbf{0.7  (0.72)} & 0.6  (0.66) \\ \hline
Interview & 0.13  (0.31) & 0.39  (0.48) & 0.44  (0.49) & 0.31  (0.38) & 0.35  (0.41) & 0.47  (0.49) & \textbf{0.62  (0.64)} & \textbf{0.55  (0.55)} & 0.41  (0.47) \\ \hline
Panel & 0.08  (0.15) & 0.3  (0.31) & 0.26  (0.31) & 0.25  (0.3) & 0.26  (0.33) & 0.29  (0.3) & \textbf{0.44  (0.46)} & \textbf{0.41  (0.43)} & 0.29  (0.32) \\ \hline
Committee & 0.03  (0.18) & 0.16  (0.24) & 0.19  (0.19) & 0.1  (0.18) & 0.14  (0.15) & 0.12  (0.16) & \textbf{0.44  (0.47)} & \textbf{0.36  (0.39)} & 0.19  (0.24) \\ \hline
Final & 0.02  (0.2) & 0.09  (0.18) & 0.11  (0.18) & 0.01  (0.07) & 0.06  (0.14) & 0.05  (0.12) & \textbf{0.38  (0.43)} & \textbf{0.22  (0.23)} & 0.1  (0.17) \\ \hline
\textbf{MEAN} & 0.12  (0.25) & 0.31  (0.38) & 0.33  (0.38) & 0.23  (0.3) & 0.28  (0.34) & 0.31  (0.35) & \textbf{0.54  (0.57)} & \textbf{0.45  (0.46)} &  \\ \hline
%\end{tabular}
\end{tabu}
\end{table*}

We see that (1) during the conversion phase, as the number of applicants is higher, all the algorithms can achieve good results. Ensemble algorithms such as XB can perform well especially due to large sample size and also the low dimensionality of the dataset. Even more complex algorithms such as NN-DO can prevent overfitting. However, during the Evaluation phase, there is a drastic drop in the number of applicants, reducing the sample size.  Also, the dimensionality of the data increases significantly as the unstructured data from text and video is added to the dataset. Therefore, (2) in the evaluation stages, the performance of the standard algorithms (SVM, XB, TKL) drop, especially due to the complexities added to the dataset. The methods based on neural networks (NN-DO, TAT) that should capture these new complexities don't have enough samples to generalize well and overfit the smaller datasets. Even NN structures with strong regularization (NN-DO) are not able to achieve good performance, because all the weights are trained in a single stage. 

We also notice that (3) algorithms based on domain adaption approaches (TKL, TAT) do not perform well, as the knowledge passed from the previous stage is not enough to reduce the overfitting problem. TKL has a better performance than TAT due to it is a simpler structure. The results from MSGTL-DA are also lower than the other versions, which shows that the domain adaptation layer introduced increases the complexity of the final NN and hurts the performance in the evaluation phase. This suggests that the discrepancy between the feature spaces cannot be reduced by the proposed approach. 

Finally, (4) our models MSGTL and MSGTL-R can outperform all other methods due to the aggressive regularization introduced by the weights from previous states. MSGTL performs slightly better than MSGTL-R, which suggests that the combination of weight transferring with the aggressive dropout can cause the model to underfit in some cases. Comparing all DA methods, we can see that the transfer weight introduced in MSGTL-DA can make the model achieve better results than TKL and TAT. Finally, MSGTL is able to improve the average results in the evaluation phase by 0.17 points (0.33$\rightarrow$0.54, 63\%) comparing to the second best method.

In Figure \ref{fig:results_lineplot}, we plot the results for each method in the four cases in the longitudinal experiment. We can see the general pattern where the algorithms perform well during the conversion phase but are not able to keep up performance in the evaluation phase. By using weights transferred from other stages, MSGTL methods can reduce the drop in performance in achieving better results in later stages.  

%--------------------------------------------------------------------------
 \subsection{Cross-validation}
%---------------------------------------------------------------------------------------
Table \ref{tab:stage_results} also shows the results for all experiments using the cross-validation setup in parenthesis in each cell. As expected, all algorithms perform better in this scenario because the data comes from the same year, reducing the chance of covariate shift between the distribution in training and test and validation sets. 

Comparing the methods, the results are similar to the longitudinal case, and the proposed method MSGTL is able to achieve the highest score in both conversion and evaluation phases. However, the gain in the cross-validation setup is 0.16 (0.38$\rightarrow$0.54, 42\%) higher than the second place, which is lower than what was achieved in the longitudinal setup, 0.17. This suggests that our method can achieve better gains in case of a covariate shift comparing to other methods.
%---------------------------------------------------------------------------------------
 \subsection{Hyperparameter Sensitivity}
 \label{sec:parameter_sensitivity}
\begin{figure}[tbp]
\centerline{\includegraphics[width=.9\columnwidth]{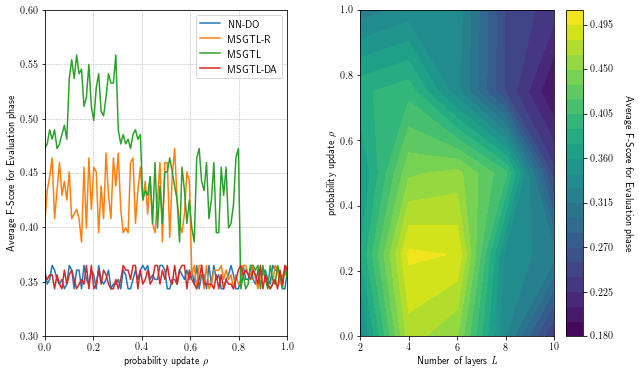}}
\caption{Results for hyperparameter sensitivity analysis. Left shows the performance with different values for $\rho$. Right shows results from changes in $\rho$ and the number of layers $L$, defined by $\omega$ and $\gamma$.}
\label{fig:parameters}
\end{figure}

%---------------------------------------------------------------------------------------
\textbf{Nodes in the final layer $\gamma$ and max number of layers $\omega$}: These two hyperparameters together control the shape of a neural network. Our proposed NN used a simple approach where each following layer has half of the nodes of the previous one until the last layer with $\gamma$ number of nodes. Since the first layer has $n$ nodes from the input features in $X$, if $n$ is significantly large, we have a structure with many layers. We suspect that in such a structure, the effect of the transfer weights from previous stages is decreased while it also becomes more complex, making it more susceptible to overfitting. This assumption is verified in Figure ~\ref{fig:parameters}, where we can see a decrease in performance as the number of layers increase.
%---------------------------------------------------------------------------------------
%---------------------------------------------------------------------------------------

\textbf{Probability of update $\rho$}: This hyperparameter controls the probabilities of updates from previous NNs. In the \textit{initialization configuration}, $\rho=1$ and $P$ contains masks with all values equal to 1. The weights from previous NNs are only used as initialization. The results in Figure ~\ref{fig:parameters} show that the NN converges to similar weights as if they were initialized randomly, performing similarly to a standard neural network. That shows that just initialization does not guarantee knowledge transfer from previous stages. For $\rho=0$, $P$ contains masks with all values equal to 0. In this \textit{fixed stage}, the knowledge transfer is guaranteed and the results are improved. However, there is a possibility of partial underfitting since many weights updates are restricted. Hence, setting $\rho$ between the two extremes achieved the best results. From many experiments, the best value is at $\rho=0.3$. In this case, knowledge is still transferred from the previous stage, but the NN can better fine-tune to the new data without overfitting. 
%-----------------------------------------
\section{Conclusion}
\label{sec:conclusion}
%---------------------------------------------------------------------------------------
In this paper, we presented a \textit{Multi-Stage Transfer Learning (MSGTL)} approach to enhance the transferability of knowledge from different neural network (NN) structures. Each NN is trained in sequential stages in a multi-stage process. The method consists of transfer knowledge from a stage with many samples and low-dimensionality to a stage with high-dimensionality but much fewer samples. We confirm that previous weights are more general and they help more complex NNs to learn specific weights in the later stages of process. We perform an empirical evaluation using a dataset from 3 years with a dual funnel structure. We compare the results with other state-of-the-art algorithms for domain adaptation and regularization and the results show that our approach was able to achieve significant improvements, especially in later stages. We also show that allowing some level of updates in the transferred weights using a probabilistic hyperparameter improves the performance in our method. Future work includes investigating the approach in other multi-stage situations in different domains and processes with a small sample size.

Selection processes are a sensitive topic. It is possible that training on existing outcomes of such processes is likely to reproduce the biases exhibited by those who did the selection in the first place. We have not investigated deep enough how our method would either increase or reduce such bias. In future work, we also plan to use explainability methods to understand how the knowledge transfer mechanism affects different classes and groups.
%---------------------------------------------------
\bibliographystyle{ecai}
\bibliography{main}
\end{document}